\documentclass[11pt]{article}

\usepackage[margin=1in]{geometry}

\usepackage[T1]{fontenc}
\usepackage[utf8]{inputenc}
\usepackage{microtype}

\usepackage{amsmath}
\usepackage{amssymb}
\usepackage{amsthm}
\usepackage{bm}

\theoremstyle{plain}

\theoremstyle{definition}

\theoremstyle{remark}

\usepackage{graphicx}
\usepackage{booktabs}
\usepackage{multirow}
\usepackage{subcaption}
\usepackage{caption}
\usepackage{algorithm}
\usepackage{algpseudocode}

\usepackage{tikz}

\usepackage{natbib}
\bibpunct[, ]{(}{)}{,}{a}{}{,}

\usepackage{hyperref}
\hypersetup{
    colorlinks   = true,
    linkcolor    = blue!60!black,
    citecolor    = blue!60!black,
    urlcolor     = blue!60!black,
    pdftitle     = {LLM-Augmented Digital Twin for Policy Evaluation in Short-Video Platforms},
    pdfkeywords  = {Short-video platform, Digital twin, Large language models, AI agents}
}
\usepackage{cleveref}

\usepackage{xcolor}
\usepackage{enumitem}
\usepackage{setspace}
\usepackage{fancyhdr}
\title{%
    \LARGE\bfseries
    LLM-Augmented Digital Twin for Policy Evaluation\\[4pt]
    in Short-Video Platforms
}

\author{%
    Haoting Zhang\\
    \small University of California, Berkeley\\
    \small \texttt{haoting\_zhang@berkeley.edu}
    \and
    Yunduan Lin\\
    \small The Chinese University of Hong Kong\\
    \small \texttt{yunduanlin@cuhk.edu.hk}
    \and
    Jinghai He\\
    \small University of California, Berkeley\\
    \small \texttt{jinghai\_he@berkeley.edu}
    \and
    Denglin Jiang\\
    \small New York University\\
    \small \texttt{florence.jiang@nyu.edu}
    \and
    Zuo-Jun (Max) Shen\\
    \small The University of Hong Kong\\
    \small \texttt{maxshen@hku.hk}
    \and
    Zeyu Zheng\\
    \small University of California, Berkeley\\
    \small \texttt{zyzheng@berkeley.edu}
}

\date{%
    \small\today\\[6pt]
}

\begin{document}

\maketitle

\begin{abstract}
Short-video platforms are closed-loop, human-in-the-loop ecosystems where platform policy,
creator incentives, and user behavior co-evolve. This feedback structure makes counterfactual
policy evaluation difficult in production, especially for long-horizon and distributional
outcomes. The challenge is amplified as platforms deploy AI tools that change what content
enters the system, how agents adapt, and how the platform operates. We propose a large
language model (LLM)-augmented digital twin for short-video platforms, with a modular
four-twin architecture (User, Content, Interaction, Platform) and an event-driven execution
layer that supports reproducible experimentation. Platform policies are implemented as
pluggable components within the Platform Twin, and LLMs are integrated as optional,
schema-constrained decision services (e.g., persona generation, content captioning, campaign
planning, trend prediction) that are routed through a unified optimizer. This design enables
scalable simulations that preserve closed-loop dynamics while allowing selective LLM
adoption, enabling the study of platform policies, including AI-enabled policies, under
realistic feedback and constraints.
\end{abstract}


\bigskip

\section{Introduction}

Short-video platforms (e.g., TikTok, Instagram Reels, and Kuaishou) have become a dominant form of online media, reaching billions of users worldwide. For instance, Instagram reported 3 billion monthly active users in September 2025 \citep{cnbc2025instagram3b}; TikTok has reached 1.9 billion monthly active users globally in 2026 \citep{demandsage2026tiktok}; and Kuaishou reported 735.6~million average monthly active users and 401~million average daily active users in the fourth quarter of 2024 \citep{prnewswire2024kuaishoufy}.

Beyond scale, short-video platforms are structurally complex: platform policies interact with strategic creator and user behavior, and the resulting co-adaptation reshapes system-level outcomes. As a result, a central challenge is \emph{policy evaluation}, that is, predicting how a proposed policy change will affect aggregate metrics when agents and algorithms adjust in response. Rigorous evaluation through production experimentation is often infeasible or inconclusive because of deployment risk, engineering overhead, and confounding from concurrent product changes. Moreover, many interventions are ethically sensitive: changes in ranking, promotion, or moderation can introduce unfairness in exposure, behavior, and downstream social outcomes at scale. These difficulties are further amplified in the AI era \citep{zhang2024daily,zhang2024does,he2025reinforcement}. As platforms delegate parts of their decision-making to AI modules (e.g., creator assistants and trend prediction), feedback loops can accelerate and become harder to audit, making AI-enabled policy interventions more difficult to test and govern \citep{zhang2024enhancing,zhang2021neural}. Together, these constraints motivate a framework for safe, controlled counterfactual experiments that respects endogenous platform dynamics. 

In short-video settings, such a framework must capture two coupled features: the platform's \emph{closed-loop} feedback and the \emph{human-in-the-loop} strategic adaptation. The system is \emph{closed-loop} because exposure shapes behavior, behavior becomes the metrics that drive subsequent exposure, and this circular dependence makes policy changes hard to evaluate while holding other factors fixed. As a result, causal attribution is challenging and long-horizon online experiments can be difficult to interpret, especially when key algorithmic mechanisms remain partially opaque to external researchers \citep{gao2023echo,hornik2025leveraging,wu2023causal}. This motivates a \emph{digital twin}, that is, a virtual replica that preserves the same feedback structure, for evaluation of specific policy modules \citep{rossetti2024social}. At the same time, the system is \emph{human-in-the-loop}: agents respond strategically to recommendation incentives and performance feedback \citep{he2025collaborative}. Compared with engineered, process-centric digital twins (e.g., manufacturing, logistics, and energy systems), short-video ecosystems are high-frequency and shaped by subjective preferences and social context, making hand-crafted rules or simple statistical response models brittle. Recent progress in large language models (LLMs) offers a promising alternative, providing flexible, context-conditioned reasoning that can generate coherent decisions and artifacts that better approximate human deliberation.

Despite these motivations, existing simulators still face a scale-fidelity trade-off: few combine high-throughput, closed-loop dynamics with realistic, human-in-the-loop adaptation. On the one hand, large-scale agent-based simulators support high-throughput, system-level experimentation \citep{hegselmann2002opinion,geschke2019triplefilter}. However, they often rely on simplified cognitive archetypes and thus miss strategic adaptation and semantic nuance, which may in turn bias system-level outcomes. On the other hand, LLM-based generative agents can produce coherent individual behaviors \citep{park2023generative}, yet they are typically evaluated in small sandboxes and rarely model the full closed-loop coevolution of users, content, platform decisions, and interactions.

Recent work begins to narrow this gap. For example, \citet{yang2024oasis} proposes \emph{OASIS}, a platform-scale simulator that couples large agent populations with a recommender system to reproduce macro-level phenomena such as diffusion and polarization over long horizons. However, it still falls short as a tool for policy evaluation on short-video platforms. In particular, OASIS does not explicitly represent the full closed-loop feedback path. It also lacks a modular policy interface and models key platform levers (e.g., trends, promotion, moderation) only coarsely, limiting counterfactual evaluation. Finally, it does not leverage LLMs to capture rich user and content semantics that can shape adaptive behavior under policy changes.

To overcome these limitations, we develop an \emph{LLM-augmented, agent-based digital twin} for short-video ecosystems. Our system integrates a four-twin architecture with an event bus to make the closed-loop cycle explicit and configurable for controlled counterfactual experiments. We use LLMs where semantic fidelity is essential (e.g., persona-conditioned reasoning, content synthesis, structured planning outputs), avoiding extensive hand-crafted rules while maintaining high-throughput simulation. Together, these design choices enable safe and interpretable evaluation of platform policies. This framework covers AI-driven tools and algorithmic components under realistic feedback, while preserving the adaptive behavior that drives platform dynamics.

\section{Literature Review}

\subsection{Short-Video Platform and Policy Evaluation}

Short-video platforms are closed-loop, human-in-the-loop ecosystems, which makes policy evaluation challenging because policy changes reshape the data-generating process. For example, recommender feedback can induce homogenization and amplify bias through self-reinforcing dynamics \citep{chaney2018algorithmic}; logged interactions are exposure-biased and thus require debiasing for offline evaluation \citep{schnabel2016unbiased}; and similar issues arise in industrial-scale recommender pipelines (e.g., YouTube) \citep{covington2016deep}. In addition, creators and users adapt strategically over time, so intervention effects may drift as the platform re-equilibrates \citep{nandy2021b,johari2022experimental}. Finally, recommendation systems act as gatekeepers, and operational choices can introduce informational barriers and normative constraints \citep{bozdag2013bias}. 

These challenges motivate two complementary toolkits for policy evaluation: online experimentation and offline observational evaluation from logged data. For online tests, two-sided structure and interference call for marketplace-style designs (e.g., producer-consumer randomization) \citep{zhang2020long,ye2025deep}, with switchbacks addressing temporal dependence \citep{bojinov2023design} and cluster/block randomization mitigating grouped interference and correlated outcomes \citep{candogan2024correlated}. For offline evaluation, logged bandit feedback enables debiased counterfactual estimation via inverse propensity weighting and doubly robust methods \citep{swaminathan2015counterfactual,zhang2025machine,zhang2023contextual}. In practice, staggered rollouts and capacity constraints (e.g., stockouts) further complicate identification and require phased-deployment designs and tailored estimators \citep{xiong2024optimal,chen2025b}. Our approach instead builds a digital-twin framework to enable counterfactual analysis under closed-loop feedback. As a result, we can provide a reusable, modular testbed that others can extend to study a broader class of policies and platform mechanisms beyond an one-off experimental design.

\subsection{LLM sandbox}

LLM sandboxes have emerged as a critical methodological bridge between static benchmarks and real-world deployment. Initial applications focused on \emph{social prototyping}, allowing designers to test system dynamics before launch. \cite{park2022social_simulacra} introduced \emph{social simulacra}, a technique that prompts LLMs to simulate thousands of community interactions, thereby exposing design flaws early in the development cycle. This approach evolved into \emph{generative agents}, which equip entities with long-term memory, reflection, and planning to support reliable behavior in open-ended environments inspired by \emph{The Sims} \citep{park2023generative}. The architectural patterns established in these works, including personas, memory retrieval, and observation-plan-act loops, have since become the standard for sandboxed agent simulation.

Beyond prototyping, sandboxes serve as rigorous evaluation environments for complex agentic behaviors. \emph{SOTOPIA} frameworks agents within diverse social scenarios to assess commonsense reasoning and strategic communication, revealing significant gaps in current model capabilities \citep{zhou2024sotopia}. In the functional domain, \emph{WebArena} constructs a realistic web execution environment, comprising e-commerce, forums, and development tools. This construction benchmarks long-horizon task completion, surfacing failure modes that remain invisible in simplified prompts \citep{zhou2023webarena}.

Recent research has shifted focus toward platform-scale ecosystems. \cite{yang2024oasis} introduced \emph{OASIS}, a modular simulator supporting up to one million agents, which integrates recommendation systems to reproduce macro-phenomena like information diffusion and polarization. Parallel work on \emph{CRSEC} explicitly models the emergence and enforcement of social norms, demonstrating that agent societies can autonomously reduce conflict through dynamic norm propagation \citep{ren2024crsec}. These efforts underscore that realistic social simulation requires modeling not just individual agents, but also the algorithmic and normative structures that govern their interaction.

Furthermore, sandboxes are increasingly utilized for alignment and training of LLMs and agents \citep{liu2023training_social_alignment,zhang2024enhancing}. \cite{liu2023training_social_alignment} demonstrate that simulating a society where agents receive collective feedback and iteratively revise behavior yields superior social alignment compared to static reward-based methods. Supporting these complex, distributed workflows requires robust infrastructure; frameworks like \emph{AgentScope} provide the necessary message-exchange architectures, tool integrations, and fault tolerance to scale simulations from small demos to reliable research platforms \citep{gao2024agentscope}.

\section{System Description}
In this section, we present the LLM-augmented, agent-based digital twin of a short-video social platform. The implementation builds on the \textit{OASIS} infrastructure \citep{yang2024oasis} and introduces a modular, scalable architecture for controlled experimentation and counterfactual studies.

At a high level, the digital twin comprises six core modules: four twins (User, Content, Interaction, and Platform), a cross-twin event system, and an environment orchestrator. Each twin characterizes a specific subsystem and maintains its state. The event system coordinates cross-twin information flow and records an append-only log for instrumentation and replay. The orchestrator advances simulated time and executes event handlers that update twin states and emit new events, thereby capturing the platform’s closed-loop dynamics.

The digital twin exposes a discrete control interface with \textbf{48 action types} and an event taxonomy of \textbf{23 cross-twin event types}. The action types define the admissible interventions, spanning both general social-platform operations (e.g., navigation and discovery, social-graph updates, posting/commenting primitives, group/community operations, and basic commerce) and content-specific behaviors (e.g., create/watch/skip/like/share/comment, with optional live/duet/stitch and gifting). Meanwhile, typed events standardize how state changes and signals propagate across twins, covering the main causal pathways: user/session intents (User$\rightarrow$Interaction), interaction outcomes and item updates (Interaction$\rightarrow$Content/User), and platform-side governance and instrumentation (Platform$\rightarrow$Content/Interaction/User). Together, this action-event layer provides a compact yet expressive contract for permissible state transitions and for tracing long-horizon feedback effects through the system.

Building on this event-driven foundation, we enable counterfactual policy evaluation through two abstractions: (i) \emph{policy components} within the Platform Twin and (ii) \emph{LLM decision services} callable during event handling. First, we represent platform policies as explicit, parameterized components inside the Platform Twin (e.g., recommendation, exposure-stage logic, trend tracking, and governance). A counterfactual run is instantiated by swapping or re-parameterizing only these policy components while holding the remaining world state fixed. This design makes long-horizon outcome differences attributable to the intended decision rules rather than to implicit cross-module state mutation. Second, we incorporate LLMs via a unified optimizer service instantiated by the Platform Twin, which centrally manages schema-constrained requests. In other words, the optimizer exposes a shared, platform-level interface for LLM calls, so any twin (User/Content/Interaction/Platform) can invoke these LLM-backed utilities from within its event handlers without implementing separate LLM clients. Overall, the framework enables selective, cost-governed LLM use for tasks such as persona and caption generation, creator campaign planning, and trend prediction.

In the following, we introduce the details of the four twins, the event bus and environment, as well as the LLM integration and cost-governance mechanisms, respectively.

\subsection{Four-Twin Architecture}
\label{subsec:system_architecture}

We adopt a four-twin architecture (User, Content, Interaction, and Platform) that decomposes the system into subsystems with explicit responsibilities and bounded state. This separation improves interpretability by making clear which twin generates each class of signals and outcomes. It also enables clean ablations and counterfactual analyses by allowing one twin to be replaced while holding the others fixed. Finally, it facilitates scalable implementation via modular development, targeted profiling, and iterative component upgrades.

\begin{figure}[ht!]
    \centering
    \includegraphics[width=0.9\linewidth]{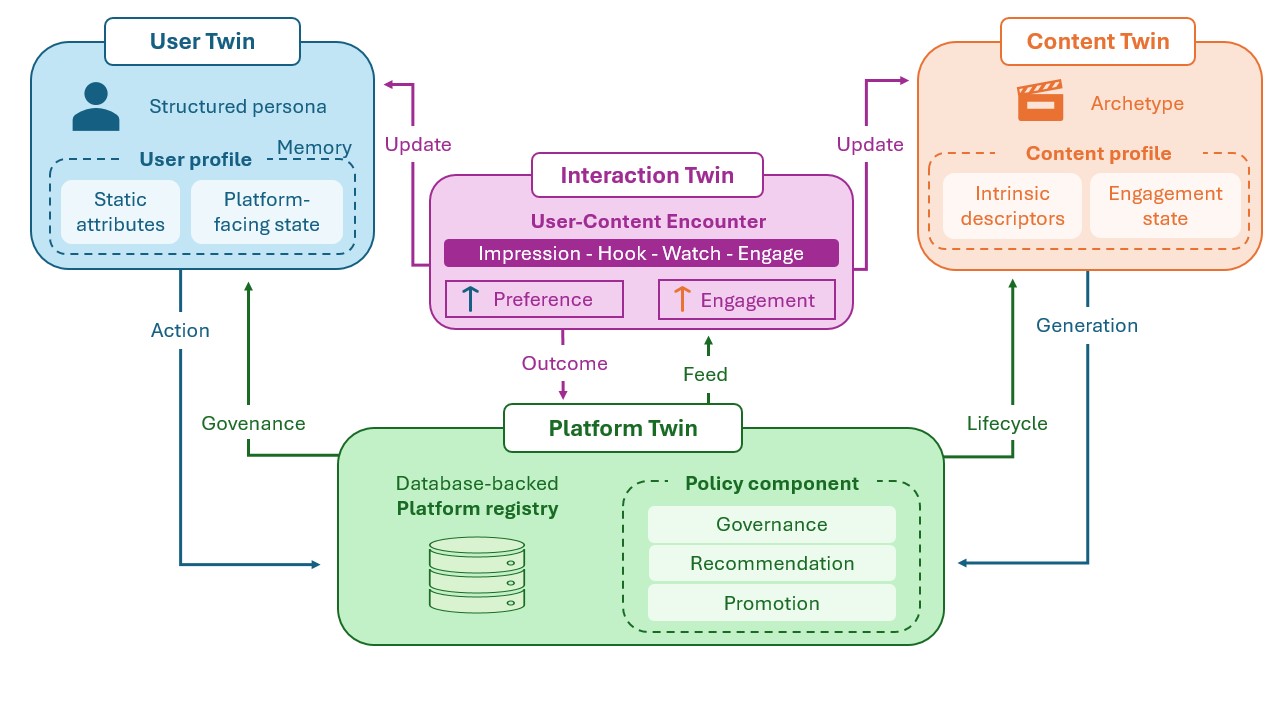}
    \caption{Illustration of Four-Twin Architecture}
    \label{fig:four_twin_architecture}
\end{figure}

Each twin comprises three components. First, it defines data structures that represent the subsystem’s state as exposed to the simulator. Second, it specifies update procedures that evolve this state in response to actions and events. Third, it interacts with the rest of the system only through a restricted interface, namely a finite action space and the typed event bus. Thus, cross-module coordination is mediated by explicit handlers rather than direct field access. Figure~\ref{fig:four_twin_architecture} summarizes the overall structure. We next describe each twin in turn and provide implementation details in Appendices \ref{sec:appendix_embeddings}-\ref{sec:appendix_platform_twin}.

\begin{itemize}
    \item \textbf{User Twin.} The User Twin models a population of autonomous user agents. Each agent is represented by a set of static attributes and an evolving, platform-facing state. The static component includes demographics, creator status, and calibrated decision propensities (e.g., attention span and like/comment/share tendencies), among other user-level descriptors. To capture creator concentration, agents are assigned to tiers (elite, active, casual, or pure consumers) drawn from a power-law distribution, so that a small fraction produces most content and attracts disproportionate attention. The evolving state primarily captures time-varying preferences, which we implement as a 50-dimensional latent preference vector.
    
    Agents are initialized from a structured persona, that is, a compact specification of identity and behavioral tendencies (e.g., background descriptors, high-level interests, and usage/engagement style). The persona is mapped to the static attributes and the initial preference state. After initialization, preferences evolve through repeated platform interactions. The User Twin maintains a memory mechanism with short- and long-term stores: recent experiences are written to short-term memory and consolidated into longer-lived preferences that drift over time, thereby updating the agent’s latent preference vector. Retention is modeled as a decay process inspired by the Ebbinghaus forgetting curve, augmented with an access-count spacing effect; salient or repeatedly retrieved experiences decay more slowly and remain influential longer. Together, these mechanisms yield a compact, evolving preference state that supports long-horizon dynamics.

    The User Twin also exposes a restricted interface to other twin modules. Specifically, it takes as input a compact session context (including served content/impressions) and returns as output a user action (e.g., consume/engage/create), along with any auxiliary decision parameters required for downstream execution.

    \item \textbf{Content Twin.} The Content Twin models the platform’s evolving corpus of short videos. Instead of generating and storing pixel-level media, we represent each item with an abstract, feature-based profile to reduce cost and improve reproducibility. Mirroring the User Twin, each content profile also contains intrinsic descriptors fixed at creation and an engagement state that evolves through interactions. Static descriptors include lightweight metadata (e.g., category, duration, hashtags, and title/description) and a fixed feature vector for efficient interest matching. In the default configuration, we maintain a compact 50-dimensional content vector for lightweight scoring, while optionally retaining richer multimodal embeddings when higher-fidelity semantic matching is needed. The dynamic engagement state records interaction-driven signals such as views, engagement rates, recency, and other derived performance statistics.

    New content is produced by an archetype-driven generator that samples from a library of canonical categories (e.g., dance, comedy, education, cooking, and pets). Upon creation, the generator selects an archetype and instantiates the static portion of the profile. Intrinsic descriptors are jointly determined by archetype-level parameters and creator characteristics inherited from the User Twin, which modulate content-quality primitives such as hook strength, watch quality, and virality potential. The content is then persisted in the content store. As exposure and interactions accrue, the Content Twin updates the dynamic portion of the profile in response to typed events (e.g., watch outcomes and engagement summaries), thereby tracking the item’s evolving engagement state.

    Accordingly, the Content Twin consumes structured events from other twins that capture new-content creation, user interaction outcomes, and platform-level status signals (e.g., trending or viral flags). It also provides read-only queries that allow the Platform Twin to retrieve eligible candidates and their features for exposure decisions, which determine what is shown and how items are scored. Finally, it emits content-state update events so other components can refresh logged metrics and policy variables without directly modifying the Content Twin’s internal state.

    \item \textbf{Interaction Twin.}
    The Interaction Twin serves as the platform’s micro-level behavior engine, operating at the resolution of a single user-content encounter. Its purpose is to convert a served impression into realized behavioral outcomes and expose them to the rest of the system as typed interaction events. To preserve modularity, the Interaction Twin keeps only lightweight, encounter-specific state: calibrated behavioral parameters such as hook thresholds, completion priors, and noise scales. User preferences and memory remain owned by the User Twin, while content descriptors and engagement aggregates remain owned by the Content Twin. The Interaction Twin consumes these profiles as inputs but does not duplicate them internally.

    For each encounter, the Interaction Twin runs an event-level simulator that emulates the short-video scrolling loop and maps profiles to outcomes (Figure~\ref{fig:interaction_loop}). The simulator proceeds in stages. First, it evaluates an \emph{immediate hook response} within a brief initial window: if the content’s hook strength falls below a calibrated threshold, the user quickly swipes away, yielding an early skip and near-zero watch time. Conditioning on passing this hook window, the simulator computes an \emph{interest-match} score between the user’s current preference representation and the content’s descriptors. When multimodal embeddings are available, this score fuses visual, audio, and textual components with explicit weights; otherwise, it falls back to cosine similarity between compact 50-dimensional user and content vectors. The resulting match score determines expected completion and downstream engagement (e.g., like, share, comment) propensities. Next, the simulator realizes \emph{watch time} by sampling a stochastic completion outcome that combines a calibrated base completion prior, an interest-driven uplift, and log-normal noise to reproduce realistic dispersion. Finally, conditional on the realized watch outcome, it samples \emph{engagement actions} using calibrated propensity models.

    The Interaction Twin exposes a narrow callable interface. Conceptually, it takes as input an encounter request assembled by the orchestrator from the User Twin (user-side state) and the Content Twin (content-side state). It outputs the realized behavioral outcome of the encounter, which the orchestrator publishes as typed interaction events for downstream updates.

    \begin{figure}[ht!]
    \centering
    \includegraphics[width=0.9\linewidth]{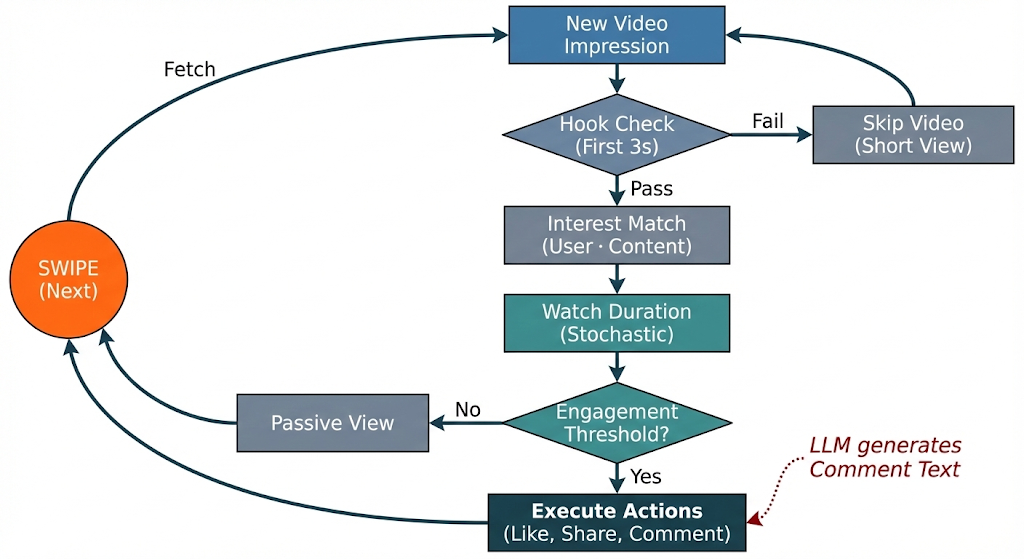}
    \caption{Interaction Twin behavioral simulation loop. The flowchart depicts the multi-phase microdynamics of a user-content encounter: a user processes a new video impression via (i) an immediate hook response, (ii) interest matching, and (iii) stochastic watch-time sampling. If engagement criteria are satisfied, engagement actions (Like, Share, Comment) are triggered, optionally invoking an LLM to generate natural-language content, before the agent swipes to continue the loop.}
    \label{fig:interaction_loop}
    \end{figure}
    
    \item \textbf{Platform Twin.} The Platform Twin encapsulates platform-side infrastructure and its policy layer. Here, \emph{platform policy} refers to configurable decision rules that map the current ecosystem state to exposure decisions and platform-side control updates---that is, what is shown, how distribution evolves over time, and which governance signals are produced. This policy layer is also the object of counterfactual experiments. In our implementation, the Platform Twin comprises three main policy components: (i) governance, (ii) recommendation, and (iii) promotion, and it can be extended with additional subsystems (e.g., moderation, ad injection, gifting, and creator monetization).

    To support these responsibilities, the Platform Twin maintains a database-backed \emph{platform registry} as the persistent system of record for platform-facing state. Although this registry is logically global, other twins may only query it through read-only accessors to obtain platform-facing features and diagnostics. It is not shared for mutation: only Platform-Twin handlers commit updates to the registry, while downstream twins update their own internal state based on the resulting typed events. We decompose the registry into two complementary parts: control state, which contains platform-adjustable parameters and registries specifying the active policy configuration and constraining admissible actions; and observational state, which consists of action/interaction logs and derived aggregates that record realized outcomes of user and platform activity. Intuitively, the control state represents the platform's policy levers (inputs to decision making), whereas the observational state captures realized ecosystem dynamics (outputs of the closed-loop system). In addition to this persistent state, the Platform Twin maintains a lightweight in-memory layer that caches rolling statistics for efficiency, while treating the database as the source of truth.

    Next, we describe the state evolution and the interfaces of the three policy components: governance, recommendation, and promotion.

    \emph{(i) Governance.} Governance is the Platform Twin’s measurement and oversight layer: it monitors ecosystem-wide activity and converts raw logs into structured diagnostics. Concretely, it maintains a trend tracker that aggregates engagement for hashtags, sounds, and categories over a rolling window and infers lifecycle signals (emergence/peak/decline), and a cascade tracker that summarizes diffusion on the social graph using descriptors such as cascade depth and branching factor. In operation, governance reads interaction logs and content metadata generated by the User, Content, and Interaction twins, together with current recommendation/promotion exposures, and returns typed governance signals that are written back into the platform state and can optionally be emitted as events for further actions.
    
    \emph{(ii) Recommendation.} Recommendation allocates exposure and forms the core feedback loop, shaping engagement and popularity. We implement several platform-inspired variants: a TikTok-style variant that mixes interest-matched, trending/viral, recent (exploration), and followed-creator candidates and ranks by relevance and performance with popularity/recency boosts; a Kuaishou-style variant that emphasizes social/community relevance and enforces creator-diversity via heavier social-connection weighting; and a hybrid variant that interpolates between the two with an optional pre-ranking exposure filter. In operation, it reads user/context signals and the candidate content pool, applies a configurable retrieve-and-rank pipeline with explicit constraints, and returns served impressions (with diagnostic traces) that drive the Interaction Twin and update platform logs.

    \emph{(iii) Promotion.} The Platform Twin implements a staged exposure pipeline for newly uploaded content. Each item is stored in a persistent stage store and begins with a small test audience (e.g., $\sim$100 impressions) to gather early signals. The platform then applies performance gates: items that meet retention/engagement thresholds advance to a larger stage (e.g., $\sim$500 impressions), while weaker items receive only limited incremental traffic. Content is promoted to a viral stage only if it sustains strong quality signals (notably high completion rates and sufficient engagement velocity). Once promoted, the platform applies an amplification factor and marks the item as broadly eligible for downstream recommendation. Operationally, promotion primarily interacts with the persistent state maintained by the Platform Twin.

\end{itemize}

\subsection{Event System and Environment Orchestrator}
\label{subsec:workflow}

To integrate the four twins and simulate their coupled evolution, we adopt a unified execution layer that couples an action-driven pathway for platform-facing requests with an event-driven pathway that propagates typed signals across twins. Both pathways are coordinated by an environment orchestrator and a shared typed event bus.

In the action-driven pathway, users in the User Twin submit actions from a finite action space (e.g., \textsc{CREATE\_VIDEO}, \textsc{WATCH\_VIDEO}) along with structured payloads. The environment orchestrator schedules these actions and routes each to the appropriate Platform-Twin action handler. Action handlers implement platform workflows by parsing inputs, retrieving required state, invoking domain logic (e.g., retrieval/ranking, content instantiation, accounting), and committing platform-side updates to the platform registry.

In the event-driven pathway, the system emits typed events to record realized outcomes and trigger cross-twin updates. Events may be produced as outputs of action handlers, by scheduled platform routines (e.g., periodic trend and exposure-stage evaluation), or by other handlers as derived signals. Each twin registers subscriptions to the event types relevant to its responsibilities; upon publication, the bus appends each event to an event log and deterministically dispatches it to subscribed handlers (e.g., via a fixed per-event-type priority order). Event handling is strictly local: each event handler may update only its own twin’s internal state (and, for the Platform Twin, its own platform registry), ensuring that no module directly mutates another twin’s fields.

Together, these pathways make coordination explicit and replayable: actions trigger platform-side execution via orchestrator-invoked handlers, while events propagate the resulting signals through the shared bus to drive downstream, twin-local updates. With these execution semantics in place, we next describe a canonical workflow: content consumption.

\paragraph{Workflow: Content consumption.}
\begin{itemize}
    \item \textbf{Action emission (User Twin).}
    A user initiates a consumption cycle by selecting actions from the finite action space, most centrally \textsc{WATCH\_VIDEO} and optional engagement actions such as \textsc{LIKE\_VIDEO}, \textsc{SHARE\_VIDEO}, and \textsc{COMMENT\_VIDEO}.

    \item \textbf{Action routing and handler execution (environment orchestrator $\rightarrow$ Platform Twin).}
    The environment orchestrator routes each submitted action to the corresponding Platform-side action handler (e.g., \texttt{watch\_video}, \texttt{like\_video}, \texttt{share\_video}, \texttt{comment\_video}) and executes it.
    Within \texttt{watch\_video}, the handler loads the required persisted content descriptors, realizes watch outcomes using the watch-behavior simulator, and then commits per-impression logs and content-level aggregates and counters.

    \item \textbf{Typed outcome events (Platform handlers $\rightarrow$ event bus).}
    After a successful commit, handlers may publish typed outcome events to the shared bus to externalize what happened for downstream bookkeeping and instrumentation.
    Concretely, \texttt{watch\_video} emits \textsc{VIDEO\_WATCHED} (or \textsc{VIDEO\_SKIPPED}); engagement handlers emit \textsc{VIDEO\_ENGAGED} together with an \texttt{engagement\_type} field (like/comment/share).

    \item \textbf{Deterministic dispatch to subscribers (event bus $\rightarrow$ environment callbacks).}
    The event bus appends published events and dispatches them to all registered subscribers. In the current implementation, environment-level callbacks subscribed to a small set of outcome events (e.g., \textsc{VIDEO\_WATCHED}, \textsc{VIDEO\_ENGAGED}) update runtime metrics and tracking summaries. 
\end{itemize}

The simulation runs in asynchronous cycles to support high concurrency, allowing many users to act simultaneously while event dispatch and database operations overlap.

\subsection{Large Language Model Integration and Cost Governance}
\label{subsec:llm_integration}

    A distinguishing feature of the system is its selective integration of LLM services. Rather than replacing the simulator’s core dynamics, LLMs are used only where semantic realism or structured reasoning is needed. 

    To make this integration scalable and cost-disciplined, the runtime enforces a three-tier execution stack: \textbf{live}, \textbf{cached}, and \textbf{surrogate}, with budget-aware routing:
    \begin{itemize}
        \item \textbf{Live tier (LLM call).} When budget permits, and a live client is configured, the optimizer issues an API call to the designated model using schema-constrained prompts and structured outputs.
        \item \textbf{Cached tier (replayable outputs).} If an identical or compatible request has been served before, the system returns a previously validated result from disk-backed JSON caches, ensuring reproducibility and amortizing cost across runs.
        \item \textbf{Surrogate tier (deterministic fallback).} When live access is disabled or budget-gated, the system falls back to deterministic rule-based or calibrated surrogate generators that preserve the same output schema, so downstream handlers never block, and the execution interface remains well-defined.
    \end{itemize}

    A dedicated optimization module orchestrates task routing, batching, caching, and progressive degradation. A budget tracker monitors spend by task and tier and triggers progressive degradation as utilization crosses configured thresholds, automatically routing requests from live $\rightarrow$ cached $\rightarrow$ surrogate when needed. When live calls are enabled, requests can also be \emph{batched} (e.g., up to 50 per batch with a short flush window) to improve efficiency under concurrency, without changing the calling semantics. Finally, the simulator supports hybrid runs in which only a targeted subset of agents or modules is routed to the live tier, while the rest remain cached/surrogate, enabling high-fidelity studies on selected cohorts without sacrificing overall throughput.

    The system exposes a task taxonomy that specifies where LLMs may be invoked and which execution tiers are permitted. Please refer to Appendix \ref{sec:appendix_llm_prompts} for implementation details and prompts.
    \begin{itemize}
        \item \textbf{User Twin initialization (persona generation).} When enabled, a live LLM can synthesize structured persona profiles (e.g., backstory, interests, engagement style), which are then mapped into agent-profile parameters to induce realistic heterogeneity. Persona outputs are cached and reused across runs; when live access is unavailable, or the budget is exceeded, the system falls back to a surrogate/mock persona generator.

        \item \textbf{Content Twin creation (caption generation).} During video creation, the content pipeline can call an LLM to generate platform-realistic titles, descriptions, and hashtags conditioned on the video archetype and current trend context. Results are cached to amortize cost; if disabled or if the call fails, the system deterministically falls back to template-based captions (optionally using cached caption assets).

        \item \textbf{Interaction Twin (engagement comment generation).} Comment text is generated via a surrogate template system by default (surrogate-only in the standard configuration), ensuring stable large-scale simulation without live-call overhead.

        \item \textbf{User-side decision policy (LLM-based action selection).} For agents configured with an LLM policy, action selection is performed via an LLM-backed agent interface (e.g., ChatAgent). This pathway is separate from the platform optimizer tiers; it can be selectively enabled for a subset of agents (with rule-based fallback for the rest).

        \item \textbf{Platform-side analysis (trend prediction and creator campaign planning).} The Platform Twin can optionally invoke a live LLM to interpret aggregate telemetry and trend states and generate structured multi-day creator campaign plans conditioned on trends and creator performance. These calls are invoked sparingly, may be cached (e.g., campaign plans), and degrade to heuristic/surrogate planners when live access is disabled or constrained.
    \end{itemize}

In summary, the system combines a modular four-twin architecture with an action--event execution layer to reproduce the closed-loop dynamics of short-video platforms. LLMs are integrated selectively and are governed by an explicit live/cached/surrogate tier with budget-aware routing, caching, and batching, thereby preserving scalability without altering the surrounding execution semantics.

\section{Experiments and Evaluation}
\label{sec:experiments}

In this section, we evaluate our proposed digital twin as a testbed for counterfactual policy evaluation in short-video platforms, with a particular emphasis on AI-enabled policies. Our goal is twofold: to assess whether the simulator reproduces qualitative patterns central to platform practice, and to evaluate how LLM-enabled decision modules behave when deployed within the platform under explicit cost constraints. 

Concretely, we organize the evaluation around two platform-managed LLM surfaces that directly enter the Platform Twin’s closed-loop control problem: \textbf{creator campaign planning}, which shapes what content enters the system and how creators adapt to feedback, and \textbf{trend prediction}, which shapes how the platform anticipates and reacts to emerging topics. The analysis proceeds through two experimental suites:
\begin{itemize}
    \item \textbf{Experiment Set 1: Creator Campaign Planning.} We study how LLM-generated short-horizon creator plans reshape attention allocation and monetization outcomes in the creator economy; see Section~\ref{sec:exp_set1}.
    \item \textbf{Experiment Set 2: Trend Forecasting and Platform Control.} We evaluate governance efficacy when the Platform Twin runs a cost-aware control loop that can optionally call an LLM-based trend predictor from platform telemetry, and we stress-test the stability of the degradation mechanism when LLM-based trend prediction turns into heuristic fallback; see Section~\ref{sec:exp_set2}.
\end{itemize}
To verify modularity at the micro level, we also include targeted ablations in Appendix~\ref{sec:exp_ablation} on auxiliary LLM touchpoints (persona generation and caption synthesis), which affect semantic realism but are not the primary policy objects in the closed-loop control layer.

\paragraph{Experimental Setup.}

All experiments are run under a scaled simulation regime to enable extensive ablation and counterfactual sweeps. Unless otherwise noted, each condition is repeated over three random seeds, and we report the mean and standard deviation across seeds. Simulation horizons are fixed at 350 steps for Experiment Set~1 and 200 steps for Experiment Set~2; we additionally run extended-horizon validations when testing long-run stability of the control loop and degradation behavior.
\paragraph{LLM Role.}
LLMs function as decision units for campaign planning and trend forecasting. To simulate production constraints, we enforce the budget-aware orchestration described in Section~\ref{subsec:llm_integration}. Usage is monitored via total spend (USD) and task-specific breakdowns. Validation runs explicitly test the triggering of surrogate fallbacks and progressive degradation under budget saturation.

\subsection{Experiment Set 1: LLM Creator Campaign Planning}
\label{sec:exp_set1}

Short-video platforms are increasingly shaped by the creator economy. As a result, creators’ decision quality, what to post, how to package it (captions/hashtags), and when to publish amid fast trend cycles, has become central to performance. Yet the ecosystem is highly complex and tightly coupled with recommendation and promotion loops, making it difficult for individual creators (or even the platform) to reliably anticipate and seize emerging trends. Meanwhile, AI tools have become mainstream in creator workflows: TikTok’s Symphony Assistant provides creative insights, script generation/refinement, and recommendations. YouTube has also introduced AI-native tooling for Shorts (e.g., Dream Screen for video ideas), lowering the cost of adapting content to audiences and trend shifts. Against this backdrop, our experiment asks: when AI guidance becomes available and adopted by some creators, does it improve engagement and monetization without increasing concentration or destabilizing trend dynamics through correlated ``best-practice'' behavior?

\subsubsection{Design.}
At a high level, this suite introduces a \emph{creator campaign planner} as a platform-side decision layer that operates upstream of content creation. For each participating creator, the Platform Twin invokes a planning interface (\textsc{CREATOR\_CAMPAIGN}) that generates a three-day roadmap specifying what the creator should post over these days, including a theme/category, a hashtag bundle, a short caption, an optional live-slot suggestion, and a call-to-action (CTA; i.e., an intended viewer prompt such as follow, comment, or join live). Note that, the resulting plan does not replace the Content Twin generator; instead, it guides downstream creation by mapping planned themes, hashtags, and captions to content-archetype choices and metadata conditioning.

\emph{LLM realization.} When LLM planning is enabled, the platform assembles a structured prompt containing the creator profile, a trend snapshot, and recent performance metrics, and dispatches a \textsc{CREATOR\_CAMPAIGN} request to the LLM. The response is a JSON payload with exactly three daily entries and fields \texttt{day\_offset, category, theme, hashtags, short\_caption, live\_slot, cta}. To improve robustness and reproducibility, the payload is normalized to a canonical schema and cached (in-memory and disk-backed), allowing repeated runs to reuse identical plans without additional LLM cost.

\emph{Heuristic realization.} When LLM planning is not enabled for a creator, the system uses a deterministic three-day campaign template as a baseline planner. The template returns exactly three daily entries in the same structured format as the LLM planner. The three days follow a fixed progression: day~0 emphasizes initial discovery (launch/awareness), day~1 emphasizes interaction (engagement-oriented follow-up), and day~2 emphasizes monetization-oriented conversion (a stronger call-to-action). This provides a stable, reproducible plan that steers content creation without conditioning on trend snapshots or recent creator performance.

We evaluate creator campaign planning via a controlled set of counterfactual experiments that would be difficult to run on a live short-video platform. Within the digital twin, we vary three factors that capture the key real-world levers shaping creator outcomes: (i) the strategy of planning, (ii) the extent of adoption, and (iii) the monetization pathways for converting attention into revenue.
\begin{itemize}
    \item \textbf{Planning strategy (S).}
    \textbf{S0} uses the deterministic heuristic planner.
    \textbf{S1} enables the GPT-4 campaign planner under the system’s gating and budget constraints. This factor isolates the incremental value of LLM-driven planning quality holding the platform environment constant.
    
    \item \textbf{Adoption rate (A).}
    We vary the fraction of creators assigned to \textbf{S1} (\(0\%, 20\%, 50\%, 100\%\)) while all remaining creators use \textbf{S0}. 
    This factor is important because creators compete for finite attention and trend exposure: partial adoption allows us to quantify both the direct gains to adopters and the ecosystem spillovers on non-adopters induced by attention reallocation and feedback through platform signals.
    
    \item \textbf{Monetization stack (M).}
    \textit{Basic} enables baseline monetization channels (e.g., ads/gifts/live as configured) while disabling commerce.
    \textit{Full Stack} additionally enables commerce, i.e., an in-platform shopping channel with product catalog and purchase flows.
    This factor matters because the economic objective of planning changes with available revenue mechanisms: campaign plans may have different value when attention can be converted via purchases in addition to engagement-based monetization.
\end{itemize}

This yields \(2 \times 4 \times 2 = 16\) experimental conditions. Each condition is repeated over three random seeds (48 total runs), and we report the mean and standard deviation across seeds.

\subsubsection{Results.} To interpret the experiments, we first report ecosystem-level averages that capture the net effect of campaign-planning quality, and then examine how the gains distribute across creators and adoption regimes. Table~\ref{tab:set1_strategy} summarizes outcomes averaged across all adoption rates and monetization regimes (48 total runs), including average watch time (seconds), view inequality (view Gini), total gift revenue, gift inequality (gift Gini), and LLM cost (USD).
\begin{table}[ht!]
    \centering
    \caption{Creator-economy outcomes aggregated across all adoption rates and monetization regimes.
    Values are reported as mean$\pm$std.}
    \label{tab:set1_strategy}
    \begin{tabular}{@{}lccccc@{}}
        \toprule
        \multirow{2}{*}{\textbf{Strategy}} &
        \multicolumn{2}{c}{\textbf{Watch and View}} &
        \multicolumn{2}{c}{\textbf{Gifts}} &
        \multirow{2}{*}{\textbf{LLM Cost (\$)}} \\
        \cmidrule(lr){2-3}\cmidrule(lr){4-5}
        & Time (s) & Gini & Revenue & Gini & \\
        \midrule
        Heuristic (S0)   & 9.680$\pm$0.114 & 0.953$\pm$0.008 & 5491$\pm$353 & 0.624$\pm$0.058 & 0.00$\pm$0.00 \\
        LLM Planner (S1) & 9.668$\pm$0.084 & 0.942$\pm$0.016 & 5690$\pm$484 & 0.584$\pm$0.075 & 2.28$\pm$1.83 \\
        \bottomrule
    \end{tabular}
\end{table}

From an engagement standpoint, overall watch time is essentially unchanged, suggesting that the LLM campaign planner does not materially alter viewing intensity. Moreover, watch outcomes remain highly concentrated under both strategies, with only a small decline under S1 (Watch-Gini drops from 0.953 to 0.942). This pattern implies that exposure and viewing allocation are driven less by creator-side planning and more by platform-side recommendation and promotion dynamics.

\begin{figure}[ht!]
\centering
\includegraphics[width=0.85\linewidth]{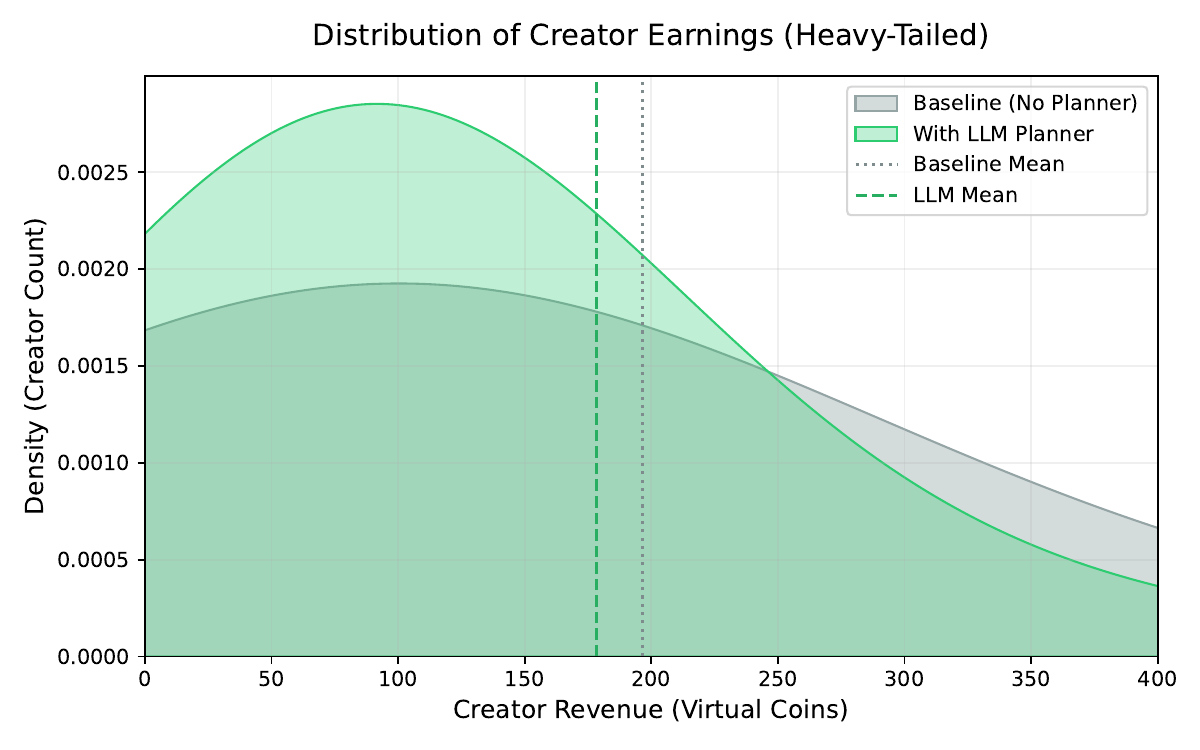}
\caption{Distribution of creator earnings. The simulation reproduces a heavy-tailed income distribution in which a small fraction of creators capture most revenue. The LLM planner (green) shifts density modestly toward the center relative to the heuristic baseline (grey), consistent with the reduction in Gift-Gini (0.62$\rightarrow$0.58).}
\label{fig:set1_revenue}
\end{figure}

In contrast, monetization responds more clearly to planning quality. The LLM planner increases mean revenue from 5491 to 5690 (+3.6\%) while simultaneously reducing revenue concentration (Gift-Gini from 0.624 to 0.584). Taken together, these results suggest that LLM planning improves creators’ monetization efficiency conditional on exposure, i.e., it increases conversion from attention into gifts without amplifying winner-take-all outcomes. Figure~\ref{fig:set1_revenue} shows the corresponding shift: relative to the heavy-tailed baseline distribution (grey), the LLM-enabled setting (green) modestly redistributes mass toward the center, consistent with reduced extreme inequality and suggesting that widely accessible planning tools can partially democratize performance.

These gains come with a modest but nonzero LLM cost (2.28 dollars on average). However, given substantial run-to-run variability, this motivates explicit cost-governance analysis in the subsequent experiment set.

\begin{figure}[ht!]
    \centering
    \includegraphics[width=0.85\linewidth]{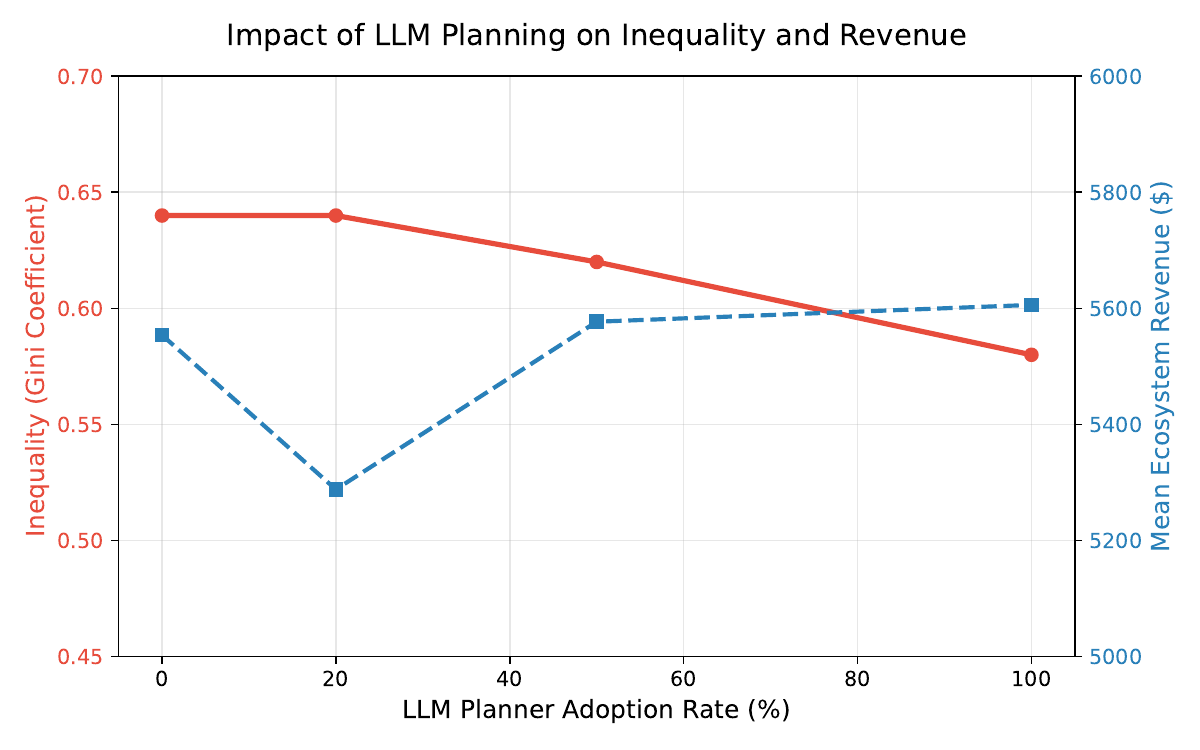}
    \caption{Ecosystem effects of LLM Planner adoption. As the adoption rate of the strategic planner increases from 0\% to 100\%, we observe a clear decrease in income inequality (Gini coefficient, red line, left axis). Total ecosystem revenue shows a modest, non-monotonic increase (blue dashed line, right axis). This suggests that widespread access to optimization tools can democratize performance without harming the overall platform economy.}
    \label{fig:set1_inequality}
\end{figure}

To further illustrate how creators’ adoption of AI tools affects outcomes, Figure~\ref{fig:set1_inequality} reports ecosystem-level results. Inequality (red line) decreases monotonically as LLM adoption scales from 0\% to 100\%, alongside a modest increase in total revenue (blue dashed line). We do not observe a stable ``early-adopter'' advantage; performance variance is mild. Additionally, the \textit{Full Stack} monetization regime yields only a small revenue increase (+94), suggesting that deeper commerce modeling (e.g., inventory dynamics) may be required for future high-fidelity studies.

\subsection{Experiment Set 2: LLM Trend Forecasting and Platform Control}
\label{sec:exp_set2}

Short-video platforms are increasingly pushed into a forecast-and-control regime, e.g., catching an emerging hashtag or sound early can yield outsized distribution and monetization gains. Meanwhile, governance failures can trigger immediate regulatory and reputational consequences, making timely detection and intervention a first-order operational requirement (e.g., Kuaishou being fined after failing to curb offensive livestream content). To address these pressures, the digital twin models the platform intervention system as a closed-loop control architecture comprising a passive sensor and an active controller. The \textbf{Trend Predictor (Sensor)} is a forecasting module that ingests live (real-time) telemetry to anticipate emerging hashtags; it serves as an information source and does not modify platform state. Conversely, the \textbf{Governance Module (Controller)} runs an asynchronous loop that reads these forecasts and executes interventions (e.g., boosting or suppression) via the platform interface. Motivated by this design, Experiment Set 2 evaluates whether LLM-based trend forecasting provides actionable early warnings to the platform, whether the explicit control loop within the platform can stabilize trend and exposure dynamics via policy interventions, and how both capabilities degrade gracefully under tight LLM budgets.

\subsubsection{Design.}
The implementation centers on a platform-side forecasting module that anticipates near-future hashtags from live telemetry. Specifically, it ingests the Platform Twin’s telemetry snapshot (e.g., engagement velocity and recency) and the trend tracker’s rolling-window state. It then outputs a structured set of candidate emerging hashtags, each with a confidence score. These predictions are written back to telemetry and consumed by downstream governance logic. In addition, we run an asynchronous control loop that, at each iteration, (i) reads telemetry (including trend forecasts and budget status), (ii) chooses governance actions subject to explicit goal constraints, and (iii) executes them via guarded wrappers that enforce safety checks and audit logging.

\emph{LLM realization.}
When the LLM predictor is enabled, the Platform Twin issues a \textsc{TREND\_PREDICTION} request through the shared LLM optimizer. A GPT-4 predictor returns a JSON list of predicted emerging hashtags together with confidence scores and brief rationales. The Platform Twin normalizes these outputs (e.g., canonical hashtag keys and confidence clamping) and persists them in telemetry so that subsequent control iterations can consume a stable, schema-consistent forecast. The platform control policy itself is rule-based instead of being decided by LLMs, while being called ``LLM-assisted'' with trends predicted by LLMs.

\emph{Heuristic realization.}
When configured as heuristic, the predictor bypasses the LLM path entirely and deterministically produces trend candidates using rule-based signals derived from telemetry. This acts as a naive forecasting baseline, ensuring that the prediction interface remains well-defined and capable of driving governance actions without incurring any LLM inference costs.

We evaluate this system via a controlled experiment comparing three distinct governance strategies ($S$) that represent increasing levels of cognitive foresight:
\begin{itemize}
    \item \textbf{S0: No Control (Baseline).} The platform operates with standard recommendation algorithms but no active governance interventions.
    \item \textbf{S1: Reactive Control (Rule-Based).} The governance module executes interventions based strictly on current telemetry thresholds (e.g., boosting a hashtag if current velocity $> 100$ posts/hour). This represents a traditional, non-predictive approach to trend management.
    \item \textbf{S2: Proactive Control (LLM-Assisted).} The governance module utilizes the \textbf{LLM Trend Predictor} (GPT-4) to anticipate viral peaks before they occur. It executes trend-aware interventions (e.g., pre-boosting predicted hits) based on the predictor's forecasts.
\end{itemize}

To evaluate cost-efficiency, we cross these strategies with three \textbf{Budget Tiers ($B$)} for the LLM optimizer: \textbf{\$100} (high), \textbf{\$50} (medium), and \textbf{\$10} (tight). This yields $3 \times 3 = 9$ experimental conditions. Each condition is repeated over three random seeds (27 total runs). In addition, we run targeted stress tests (tight budgets and aggressive control intervals) to reliably trigger budget saturation and verify that the system degrades without breaking the control interface.

\subsubsection{Results.} Table~\ref{tab:set2_governance} summarizes the experimental outcomes, including average watch time, skip rate, view inequality (View-Gini), diversity (Hashtag Entropy), and LLM cost (USD).
\begin{table}[ht!]
    \centering
    \caption{Set 2 platform outcomes by governance strategy (27 runs). Values (mean$\pm$std) are averaged across 3 budget tiers per strategy.}
    \label{tab:set2_governance}
    \begin{tabular}{@{}lccccc@{}}
    \toprule
    \textbf{Governance} & \textbf{Watch (s)} & \textbf{Skip} & \textbf{H-Entropy (bits)} & \textbf{View-Gini} & \textbf{LLM (\$)} \\
    \midrule
    None (S0)        & 9.656$\pm$0.057 & 0.363$\pm$0.005 & 4.469$\pm$0.034 & 0.886$\pm$0.023 & 2.30$\pm$0.44 \\
    Rule-based (S1)  & 9.674$\pm$0.079 & 0.363$\pm$0.004 & 4.469$\pm$0.034 & 0.893$\pm$0.023 & 2.31$\pm$0.45 \\
    LLM-assisted (S2) & 9.785$\pm$0.018 & 0.361$\pm$0.001 & 4.469$\pm$0.034 & 0.964$\pm$0.003 & 2.80$\pm$0.25 \\
    \bottomrule
    \end{tabular}
\end{table}

We observe that all metrics remain nearly unchanged between S0 and S1, whereas S2 shows clear differences. Specifically, S2 increases mean watch time from 9.656s to 9.785s and slightly reduces the skip rate. Importantly, it preserves semantic diversity (Hashtag Entropy $\approx 4.47$ bits). This stability suggests that the LLM primarily improves content quality, boosting high-grade videos within existing topics, without narrowing the platform's topical breadth (which would reduce entropy). Overall, the system achieves a Pareto improvement: higher engagement without the ``filter bubble'' effect often associated with aggressive optimization.

Moreover, LLM-driven optimization increases exposure concentration (View-Gini 0.964) relative to the baseline (0.886), highlighting a trade-off between engagement gains and inequality. Regarding cost, Table~\ref{tab:set2_governance} shows a stable baseline LLM spend ($\approx \$2.30$) driven by ongoing creator-side usage; the marginal cost of enabling the Trend Predictor is negligible ($\approx \$0.05$ per run, 2.1\% of total). This indicates that platform-side sensing is an economically efficient add-on that improves engagement without materially increasing the simulation's operational budget. With additional experiments tracking LLM costs under different budget caps, we also notice that the system remains stable across budget tiers. Even under the tightest (\$10) constraint, the degradation mechanism limits usage (up to 28.7\%) without a collapse in key metrics, confirming that budget-aware deployment is a meaningful experimental variable.

\begin{figure}[t]
    \centering
    \includegraphics[width=\linewidth]{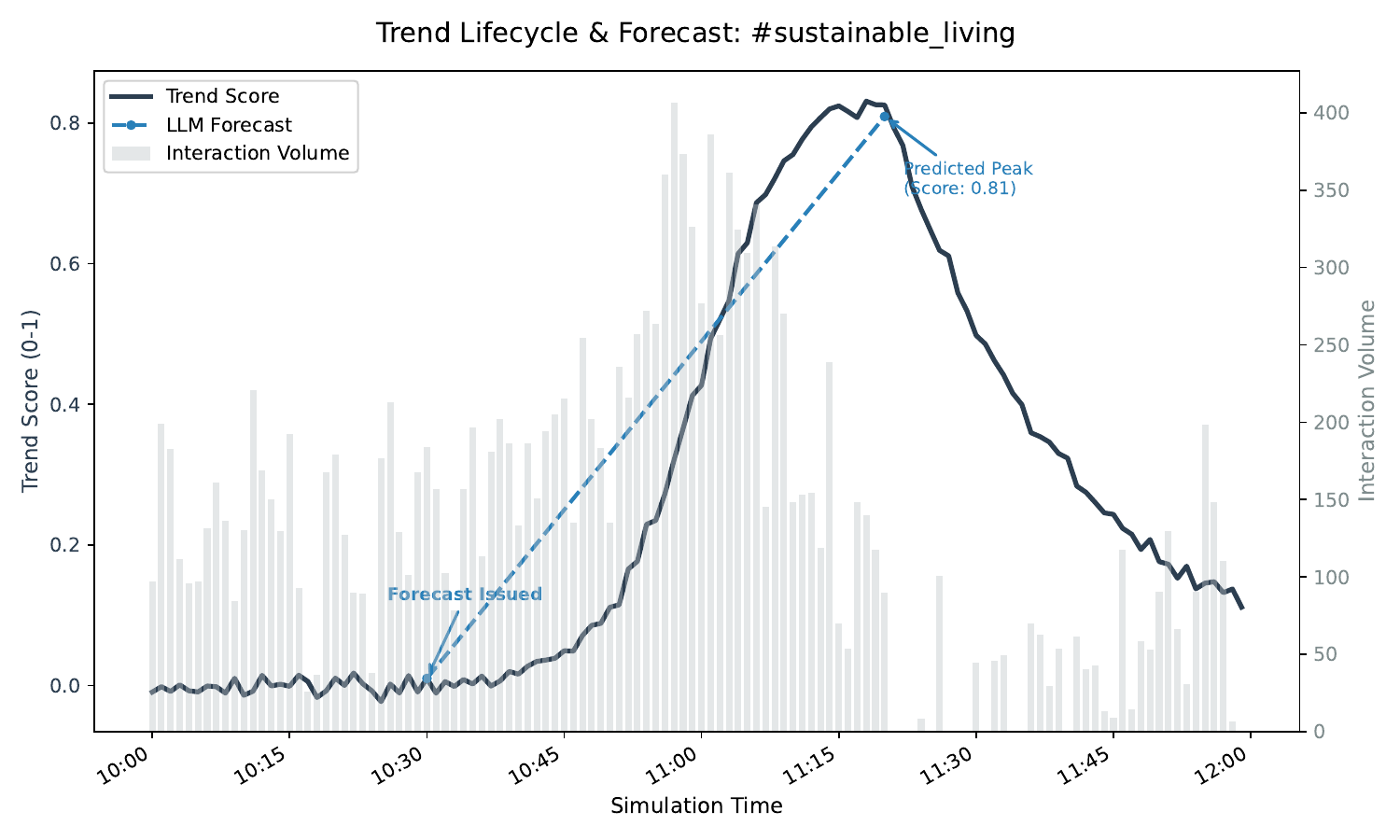}
    \caption{Trend Lifecycle and Forecasting. The solid line tracks the trend score of a viral hashtag (e.g., \#sustainable\_living) over time, driven by user interaction volume (grey bars). The dashed line represents the LLM-generated forecast issued during the emergence phase, demonstrating the Platform Twin's ability to anticipate viral peaks before they occur.}
    \label{fig:trend_lifecycle}
\end{figure}

Finally, Figure \ref{fig:trend_lifecycle} provides visual confirmation of the platform's closed-loop dynamics. We observe a distinct ``emergence'' phase in which interaction volume (grey bars) rises before the trend score spikes. Crucially, the LLM forecast (dashed line) anticipates the subsequent trajectory during the early emergence window, supporting the digital twin's role as a predictive sandbox for evaluating algorithmic interventions.

\section{Conclusion}

We develop an LLM-assisted digital twin for short-video platforms that enables controlled, reproducible policy evaluation in a closed-loop, human-in-the-loop setting. The system follows a modular four-twin architecture (User, Content, Interaction, and Platform), coupled with an action-driven execution pipeline. Exposure decisions, user responses, content evolution, and policy updates co-evolve through explicit, replayable state transitions. LLMs are used selectively as schema-constrained decision services, while a unified optimization layer governs cost through live, cached, and surrogate execution tiers. This design maintains high-throughput simulation at scale while injecting semantic realism where it is most consequential for control. Empirically, our experiments show that the twin provides diagnostic value for both platform-policy assessment and AI-enabled policy evaluation.

Several extensions remain, spanning scalability, fidelity, human-in-the-loop experimentation, and the evaluation of additional platform-relevant AI decision modules. First, we will improve scalability via parallel execution and surrogate modeling so that large-population regimes (e.g., one million users) can be simulated without changing the action/event interface or the cost-governance contract. Second, we will enhance fidelity by integrating multimodal generative models for pixel-level video synthesis and by expanding the economic subsystem. Third, we plan to implement explicit human-in-the-loop interfaces that allow researchers to interactively test policy interventions and moderation strategies against dynamic, adaptive populations. Finally, we will broaden the set of platform-relevant AI decision modules evaluated within the twin (e.g., governance workflows and operational analytics) to better support emerging needs in AI policy evaluation.

\bibliographystyle{apalike}
\bibliography{sample}

\newpage
\appendix
\section{Embedding Systems \& Vector Representations}
\label{sec:appendix_embeddings}

The digital twin utilizes a specialized, computationally efficient embedding architecture designed for high-throughput simulation. Rather than relying on heavy, real-time inference (e.g., running CLIP for every video impression), the system employs a combination of \textit{deterministic seeded embeddings} for reproducibility and \textit{evolvable low-dimensional vectors} for dynamic preference learning.

\subsection{User Agent Representation}
Agents maintain a two-tiered vector state to balance static profiling with dynamic taste evolution. The active system relies on a dense 50-dimensional vector that evolves via reinforcement learning.

\begin{table}[h]
\centering
\caption{User Agent Vector Specifications}
\label{tab:agent_vectors}
\begin{tabular}{@{}l c p{8.5cm}@{}}
\toprule
\textbf{Vector Field} & \textbf{Dims} & \textbf{Function \& Lifecycle} \\ \midrule
\texttt{content\_interests} & 50 & \textbf{Active Learner.} Initialized uniformly ($0.1$). This vector serves as the primary mutable representation of user taste, updated in real-time by the \texttt{VideoFeedbackLearner} based on watch history. \\ \midrule
\begin{tabular}[t]{@{}l@{}}\texttt{visual\_pref} \\ \texttt{audio\_pref} \\ \texttt{text\_pref}\end{tabular} & \begin{tabular}[t]{@{}c@{}}512 \\ 128 \\ 768\end{tabular} & \textbf{Architectural Slots.} Reserved for future multi-modal expansion. Currently initialized to \texttt{None}. The Interaction Twin employs a projection fallback mechanism (using \texttt{content\_interests}) when these specific modality vectors are empty. \\ \bottomrule
\end{tabular}
\end{table}

\subsection{Video Content Representation}
Content is represented by two distinct vector types: a compact vector for the recommendation engine (RecSys) and high-dimensional synthetic embeddings for the interaction simulation.

\subsubsection{Compact Content Vector (50-dim)}
The \texttt{VideoContentFeatures} vector is structured specifically for efficient similarity search in the recommender system. It is constructed deterministically from metadata:
\begin{itemize}
    \item \textbf{Dimensions 0--9 (Categorical):} One-hot encoding of the video's primary archetype (e.g., Dance, Comedy, Education, Pets).
    \item \textbf{Dimensions 10--19 (Reserved):} Reserved slots for mood/style attributes.
    \item \textbf{Dimensions 20--39 (Semantic Hashing):} Keywords extracted from captions and hashtags are hashed into these buckets, allowing text features to influence retrieval without expensive NLP inference.
    \item \textbf{Dimensions 40--49 (Stochastic):} Initialized with noise ($\mathcal{U}[0, 0.5]$) to ensure diversity in retrieval results and prevent identical content collisions.
\end{itemize}

\subsubsection{High-Dimensional Archetype Embeddings}
To simulate multi-modal matching (Visual/Audio/Text) without storing pixel data, the \texttt{VideoArchetype} system generates synthetic embeddings using \textbf{Seeded Hashing}. This ensures that a video with specific metadata (e.g., ``Archetype: Cooking'', ``Keywords: vegan, pasta'') always produces the exact same vector across simulation runs, guaranteeing reproducibility.

\begin{itemize}
    \item \textbf{Visual Embedding (512-dim):} $v \sim \mathcal{N}(0,1)$ seeded by $\text{hash}(\text{visual\_keywords} + \text{creator\_id})$. Represents visual style and quality.
    \item \textbf{Audio Embedding (128-dim):} $a \sim \mathcal{N}(0,1)$ seeded by $\text{hash}(\text{audio\_keywords})$. Represents background tracks/speech.
    \item \textbf{Caption Embedding (768-dim):} $t \sim \mathcal{N}(0,1)$ seeded by $\text{hash}(\text{title} + \text{description})$. Represents semantic content.
\end{itemize}

\subsection{Vector Interaction \& Feedback Loop}
The system links these representations through two primary mechanisms to drive the simulation loop:

\paragraph{1. The Watch Decision Logic (Fallback Projection).}
The Interaction Twin computes an \textit{Interest Match Score} by comparing agent preferences to video content. Since the high-dimensional agent slots are currently empty, the system employs a \textbf{Fallback Projection}: the 50-dimensional \texttt{content\_interests} vector is compared against truncated slices of the video's high-dimensional embeddings (e.g., \texttt{visual\_embedding[:50]}). This mathematically bridges the user's evolving low-dim preferences with the content's static high-dim attributes.

\paragraph{2. The Feedback Learner.}
The \texttt{VideoFeedbackLearner} closes the loop by modifying the agent's \texttt{content\_interests} vector after every interaction:
\begin{itemize}
    \item \textbf{Positive Reinforcement:} If $\text{watch\_time} > 80\%$ completion or $\text{action} == \text{LIKE}$, the agent's vector is nudged towards the video's \texttt{content\_vector} (Rate: $+0.3$).
    \item \textbf{Negative Reinforcement:} If the skip is immediate ($< 3$s), the vector is pushed away (Rate: $-0.2$).
\end{itemize}
\section{User Twin Specification (Agent Logic)}
\label{sec:appendix_user_twin}

The User Twin models the population of agents, managing their static personas, dynamic states, and cognitive processes. It functions as the origin point for all stochastic actions within the simulation.

\subsection{Agent Architecture}
Agents are instantiated via the \texttt{ShortVideoAgentProfile} class, which acts as a composite state container. The architecture distinguishes between innate immutables and evolving state vectors.

\begin{description}
    \item[Innate Persona (Static):] Attributes fixed at initialization time, determining the agent's baseline behavior.
    \begin{itemize}
        \item \textbf{Cognitive Traits:} \texttt{attention\_span} ($0.1-120.0$s), \texttt{humor\_affinity} ($0.0-1.0$), and \texttt{toxicity\_tolerance} (threshold for reporting).
        \item \textbf{Creator Attributes:} \texttt{creator\_tier} (Categorical: Elite, Active, Casual, Consumer) and \texttt{domain\_expertise} (mapped to specific video archetypes).
        \item \textbf{Creation Probability:} Calibrated via a power-law distribution where Elite creators produce content daily, while Consumers ($90\%$ of population) never post.
    \end{itemize}

    \item[Dynamic State (Mutable):] Attributes that evolve based on simulation interactions.
    \begin{itemize}
        \item \textbf{Social Graph:} \texttt{follower\_count}, \texttt{following\_count}, and an adjacency list for the \texttt{follower\_network}.
        \item \textbf{Session State:} \texttt{energy\_level} (decays non-linearly with watch time), \texttt{boredom\_counter} (triggers \texttt{EXIT} action), and \texttt{satisfaction\_rolling\_window}.
    \end{itemize}
    
    \item[Memory System:] The twin implements an Ebbinghaus-inspired decay model. The retention strength $R$ of a specific creator or topic at time $t$ follows $R(t) = S \cdot e^{-t/\tau}$, where $S$ is the signal strength (boosted by high-engagement actions like \texttt{SHARE}) and $\tau$ is the decay constant.
\end{description}

\subsection{Persona Generation Pipelines}
To balance scale with fidelity, the User Twin supports two initialization pipelines:
\begin{itemize}
    \item \textbf{Template-Based (Heuristic):} Fast generation using predefined archetypes (e.g., ``The Hypebeast'', ``The Critic'') for background population scaling.
    \item \textbf{LLM-Based (Generative):} High-fidelity initialization using \texttt{PersonaLLMAdapter} (GPT-4) to generate rich backstories and non-uniform interest vectors, utilized primarily for the ``Creator Economy'' experiment set.
\end{itemize}

\section{Content Twin Specification (Artifacts)}
\label{sec:appendix_content_twin}

The Content Twin manages the generation, representation, and lifecycle of video assets. It decouples the \textit{semantic} representation of a video from its \textit{pixel} data, allowing for high-throughput simulation of visual media.

\subsection{Video Archetypes}
The system defines 12 core archetypes, each with distinct statistical signatures for duration, engagement potential, and production cost.

\begin{table}[h]
\small
\centering
\caption{Content Twin Video Archetype Definitions}
\label{tab:video_archetypes}
\resizebox{\textwidth}{!}{%
\begin{tabular}{@{}l c c l@{}}
\toprule
\textbf{Archetype} & \textbf{Duration (s)} & \textbf{Viral Potential} & \textbf{Statistical Signature} \\ \midrule
\texttt{DANCE} & $15 \pm 5$ & High ($0.8$) & High hook strength, music-synced execution. \\
\texttt{COMEDY} & $45 \pm 15$ & High ($0.9$) & Narrative setup required, high completion reward. \\
\texttt{EDUCATIONAL} & $55 \pm 10$ & Medium ($0.5$) & Information-dense, low re-watchability. \\
\texttt{GAMING} & $30 \pm 10$ & Medium ($0.6$) & High visual chaos, niche audience alignment. \\
\texttt{LIFESTYLE} & $25 \pm 10$ & Low ($0.3$) & Aesthetic focus, lower hook strength. \\
\texttt{MUSIC} & $20 \pm 5$ & High ($0.7$) & Audio-dominant, high share probability. \\
\texttt{PETS} & $12 \pm 4$ & High ($0.8$) & Universal appeal, very short duration. \\
\texttt{DIY\_CRAFTS} & $50 \pm 15$ & Medium ($0.4$) & Process-oriented, satisfying visual patterns. \\
\texttt{TECH} & $40 \pm 10$ & Medium ($0.5$) & News/Reviews, text-heavy. \\
\texttt{BEAUTY} & $25 \pm 8$ & Medium ($0.6$) & Transformational content, visual-dominant. \\
\texttt{FITNESS} & $35 \pm 10$ & Medium ($0.5$) & Instructional loops. \\
\texttt{NEWS} & $55 \pm 5$ & Low ($0.3$) & Speech-heavy, requires high attention. \\ \bottomrule
\end{tabular}%
}
\end{table}

\subsection{Content Layers}
Every video object is composed of three functional layers:
\begin{enumerate}
    \item \textbf{Metadata (Descriptive):} Structured fields (\texttt{title}, \texttt{hashtags}, \texttt{category}) used for heuristic filtering and trend tracking.
    \item \textbf{Predictors (Simulation Physics):} Quantitative scores acting as ground truth for agent reactions.
    \begin{itemize}
        \item \texttt{hook\_strength} ($0-1$): Causally determines the probability of a $<3$s skip.
        \item \texttt{quality\_score} ($0-1$): Modulates watch duration and completion rate.
    \end{itemize}
    \item \textbf{Embeddings (Algorithmic):} High-dimensional vectors (Visual 512-dim, Audio 128-dim) used exclusively by the Platform Twin's recommendation engine for similarity scoring (see Appendix \ref{sec:appendix_embeddings}).
\end{enumerate}

\section{Interaction Twin Specification (Dynamics)}
\label{sec:appendix_interaction_twin}

The Interaction Twin acts as the physics engine of the simulation, resolving agent-content encounters into discrete events and routing them via a strict Event Bus.

\subsection{Action Space and Taxonomy}
The system defines 48 distinct \texttt{ActionType} members, categorized by their functional impact:

\begin{itemize}
    \item \textbf{Consumption Actions:} \texttt{WATCH\_VIDEO}, \texttt{SKIP\_VIDEO} (explicit $<3$s exit), \texttt{REFRESH} (feed reload).
    \item \textbf{Engagement Actions:} \texttt{LIKE}, \texttt{SHARE} (triggers cascade), \texttt{COMMENT}, \texttt{DUET}, \texttt{STITCH}.
    \item \textbf{Social Actions:} \texttt{FOLLOW}, \texttt{UNFOLLOW}, \texttt{SEND\_GIFT} (monetary transfer).
    \item \textbf{Navigation:} \texttt{SEARCH\_USER}, \texttt{SEARCH\_POSTS}, \texttt{EXIT}.
\end{itemize}

\subsection{Probabilistic Watch Logic}
For every impression, the \texttt{WatchBehavior} module computes the outcome based on the interaction between Agent State and Content Predictors:

\begin{equation}
    P(\text{skip}) = \sigma \left( \alpha \cdot \frac{1}{\text{hook\_strength}} + \beta \cdot \frac{1}{\text{attention\_span}} + \epsilon \right)
\end{equation}

If the video is not skipped, the \texttt{watch\_duration} is sampled from a log-normal distribution centered on the video's \texttt{expected\_completion\_rate}, modulated by the user's \texttt{interest\_match} score.

\subsection{Event Bus System}
The twin operates on a publish-subscribe model. Key event types include:
\begin{itemize}
    \item \texttt{VIDEO\_WATCHED}: Payload includes \texttt{watch\_time}, \texttt{completion\_rate}, \texttt{is\_skipped}.
    \item \texttt{VIDEO\_GOES\_VIRAL}: Emitted by the Platform Twin when velocity exceeds thresholds.
    \item \texttt{BUDGET\_EXCEEDED}: Triggers the surrogate fallback mode for LLM operations.
\end{itemize}

\section{Platform Twin Specification (System Logic)}
\label{sec:appendix_platform_twin}

The Platform Twin models the centralized infrastructure, including the recommendation engine, exposure pipelines, and the virtual economy.

\subsection{Recommendation Engine Funnel}
The recommender implements a standard industrial two-stage architecture:
\begin{enumerate}
    \item \textbf{Candidate Retrieval (Recall):} Fetches $N \approx 100$ candidates from three sources:
    \begin{itemize}
        \item \textit{Social Pool:} Content from followed creators (chronological).
        \item \textit{Viral Pool:} Top 1\% of videos by global velocity (global bandits).
        \item \textit{Semantic Pool:} Approximate Nearest Neighbor (ANN) search using the agent's \texttt{content\_interests} vector.
    \end{itemize}
    \item \textbf{Ranking (Scoring):} Candidates are scored via a weighted utility function:
\begin{align*}
    Score(u, v) = &w_1 \cdot \text{Sim}(u, v) + w_2 \cdot Q_{video} \\
   & + w_3 \cdot \text{Recency} + w_4 \cdot \text{SocialBoost}.
\end{align*}
    \item \textbf{Re-ranking:} Applies diversity filters (e.g., deduplicating creators) and ad injection logic.
\end{enumerate}

\subsection{Graduated Exposure Pipeline}
To simulate the ``cold start'' problem and viral mechanics, content passes through gated exposure states:
\begin{enumerate}
    \item \texttt{INITIAL\_EXPOSURE}: Random seed audience (approx. $200-500$ impressions).
    \item \texttt{EXPANDED\_EXPOSURE}: Triggered if engagement rate $> 15\%$. Content enters the general pool.
    \item \texttt{VIRAL\_STAGE}: Triggered if interaction velocity $> 100$ events/hour. Content is injected into the ``Viral Pool'' for global distribution.
\end{enumerate}

\subsection{Trend \& Cascade Tracking}
\begin{itemize}
    \item \textbf{TrendTracker:} Monitors hashtag velocity over rolling windows (epochs). It persists lifecycle states (Emergence, Peak, Decline) to the database.
    \item \textbf{CascadeTracker:} Maintains share trees to analyze diffusion depth and branching factors, enabling the study of information propagation dynamics.
\end{itemize}

\section{LLM Integration \& Prompt Engineering}
\label{sec:appendix_llm_prompts}

To ensure reproducibility and cost-efficiency, the digital twin utilizes a tiered LLM architecture. We employ \textbf{GPT-4-Turbo} for high-complexity reasoning tasks (e.g., strategic planning, trend forecasting) and \textbf{GPT-3.5-Turbo} or \textbf{Rule-Based Surrogates} for high-volume, low-latency tasks (e.g., commenting, basic captioning).

\subsection{Prompt Engineering Strategies}

\subsubsection{Agent Persona Generation}
\textbf{Model:} GPT-4-Turbo (Temperature: 1.0) \\
\textbf{Trigger:} Initialization of ``Elite'' and ``Active'' creator agents. \\
\textbf{System Prompt Structure:}
\begin{quote}
\textit{``You are a simulation engine for a short-video platform. Generate a detailed persona for a \{tier\} creator interested in \{domain\}. Output a JSON object containing: bio, core\_traits (5 adjectives), viewing\_preferences (vector description), and creation\_style.''}
\end{quote}
\textbf{Constraint:} Output must parse into the \texttt{ShortVideoAgentProfile} schema.

\subsubsection{Strategic Campaign Planning (Experiment Set 1)}
\textbf{Model:} GPT-4-Turbo (Temperature: 0.7) \\
\textbf{Trigger:} Daily cycle for creators participating in the monetization program. \\
\textbf{Input Context:}
\begin{itemize}
    \item \textbf{History:} Last 5 videos' performance (views, likes, retention).
    \item \textbf{Market:} Top 3 currently trending hashtags from the \texttt{TrendTracker}.
\end{itemize}
\textbf{Task:} \textit{``Analyze the performance history. Select one trending hashtag to ride and propose a video concept that bridges your niche with this trend.''}

\subsubsection{Trend Forecasting (Experiment Set 2)}
\textbf{Model:} GPT-4-Turbo (Temperature: 0.4) \\
\textbf{Trigger:} Hourly telemetry snapshot analysis. \\
\textbf{Input Context:} A time-series summary of hashtag velocity: \texttt{\{hashtag: [t-3, t-2, t-1, current]\}}. \\
\textbf{Task:} \textit{``Identify emerging trends that are in the 'Early Adopter' phase. Ignore stable or declining trends. Output a confidence score (0-1) for each prediction.''}

\subsection{Cost Management \& Surrogate Models}
Given the high volume of simulation steps (e.g., 10,000 agents $\times$ 50 interactions), direct LLM inference for every action is computationally prohibitive. We implement a \textbf{Progressive Degradation System}:

\begin{enumerate}
    \item \textbf{Priority Queue:} LLM budget is reserved for ``Elite'' agents and ``Platform Governance'' tasks.
    \item \textbf{Surrogate Fallback:}
    \begin{itemize}
        \item If \texttt{budget\_usage > 80\%}: All ``Comment Generation'' requests revert to a template library (e.g., [``Love this!'', ``Great vid'', ``lol'']).
        \item If \texttt{budget\_usage > 95\%}: ``Persona Generation'' reverts to the static \texttt{PERSONA\_TEMPLATES} dictionary.
    \end{itemize}
    \item \textbf{Caching:} All LLM responses are hashed by \texttt{(prompt, model, temp)} and cached locally to prevent redundant inference during development re-runs.
\end{enumerate}

\section{Illustrative Ablation: LLM Personas $\times$ Captions}
\label{sec:exp_ablation}

To illustrate micro-level modularity and selective LLM integration, we conduct ablations using a $2\times2$ factorial design over 200 steps. We vary whether (P) user personas and (C) video captions are produced by deterministic templates (0) or an LLM service (1). P0/P1 affects only User Twin initialization and C0/C1 affects only Content Twin initialization; all other settings are fixed.

Table~\ref{tab:ablation_session} reports session-level engagement statistics averaged over three random seeds.
\begin{table}[h]
\centering
\caption{Illustrative ablation (200-step horizon): session-level metrics. Values are mean$\pm$std across 3 seeds.}
\label{tab:ablation_session}
\begin{tabular}{@{}lcccc@{}}
\toprule
\textbf{Condition} & \textbf{Watch (s)} & \textbf{Compl.\ Rate} & \textbf{Skip Rate} & \textbf{Sess.\ Len.} \\
\midrule
P0C0 (tmpl.\ pers., tmpl.\ capt.)  & 9.75$\pm$0.11 & 0.34$\pm$0.00 & 0.36$\pm$0.00 & 21.00$\pm$0.01 \\
P0C1 (tmpl.\ pers., LLM capt.)     & 9.77$\pm$0.09 & 0.34$\pm$0.00 & 0.36$\pm$0.00 & 21.01$\pm$0.05 \\
P1C0 (LLM pers., tmpl.\ capt.)     & 8.14$\pm$0.09 & 0.28$\pm$0.00 & 0.43$\pm$0.00 & 20.99$\pm$0.03 \\
P1C1 (LLM pers., LLM capt.)        & 8.20$\pm$0.05 & 0.28$\pm$0.00 & 0.43$\pm$0.00 & 21.00$\pm$0.04 \\
\bottomrule
\end{tabular}
\end{table}
Comparing conditions reveals two patterns. First, switching from template to LLM personas yields a large, consistent shift, especially in early-session behavior: mean watch time drops from $\sim 9.76$s to $\sim 8.17$s (about $16\%$), completion decreases, skip increases, while session length is unchanged. This suggests the pipeline is sensitive to persona-conditioned behavioral parameters that propagate through interaction dynamics.

Second, caption generation has negligible marginal impact here, barely changing metrics under either persona regime. Under the current scoring/interaction setup, captions mainly affect lightweight metadata and weakly influence watch/skip outcomes unless additional semantic-conditioning channels are enabled. Thus, the baseline is primarily sensitive to user-side heterogeneity rather than caption semantics.

Overall, the ablation experiments provide a sanity check: changing only User Twin initialization (personas) produces a predictable, directionally consistent shift in session-level outcomes without modifying the rest of the simulator. The effect size also underscores a calibration need: short-horizon engagement is sensitive to persona distributions, so matching long-horizon behavior requires tuning persona-to-parameter mappings and validating aggregates against platform logs.

\end{document}